\documentclass[runningheads]{llncs}
\usepackage{graphicx}
\usepackage{amsmath,amssymb} 
\usepackage{color}
\usepackage{times}
\usepackage{epsfig}
\usepackage{tabularx}
\usepackage{subfigure} 
\usepackage{caption}
\newcommand{\mycomment}[1]{}
\newcommand{\reffig}[1]{Fig.~\ref{#1}}

\usepackage{float}
\DeclareRobustCommand\onedot{\futurelet\@let@token\@onedot}
\def\etal{\emph{et al}.}

\newcommand{\refsec}[1]{Sec.~{\ref{#1}}}

\usepackage[pagebackref=true,breaklinks=true,letterpaper=true,colorlinks,bookmarks=false]{hyperref}
\usepackage[width=122mm,left=12mm,paperwidth=146mm,height=193mm,top=12mm,paperheight=217mm]{geometry}

\begin{document}
\pagestyle{headings}
\mainmatter

\title{Fast, Exact and Multi-Scale Inference for Semantic Image Segmentation with Deep Gaussian CRFs}
\titlerunning{Fast, Exact and Multi-Scale Inference for Semantic Image Segmentation}
\captionsetup[subfigure]{labelformat=empty}
\authorrunning{Siddhartha Chandra \& Iasonas Kokkinos}
\author{Siddhartha Chandra \hfill Iasonas Kokkinos\\
{\tt \small siddhartha.chandra@inria.fr} \hfill {\tt \small iasonas.kokkinos@ecp.fr}
}
\institute{
INRIA GALEN \& Centrale Sup\'elec, Paris, France\\
}

\maketitle

\begin{abstract}
	In this work we propose a structured prediction technique that combines the virtues of Gaussian Conditional Random Fields (G-CRF)
	with Deep Learning:
	(a) our structured prediction task has a unique  global optimum that is obtained exactly from the solution of a linear system 
	(b) the gradients of our model parameters 
	are analytically computed using closed form expressions, in contrast to the memory-demanding
	contemporary deep structured prediction approaches \cite{crfrnn,Vemulapalli_2016_CVPR} that rely on back-propagation-through-time,
	(c) our pairwise terms do not have to be simple hand-crafted expressions, 
	as in the line of works building on the DenseCRF~\cite{crfrnn,deeplab1}, but can rather be `discovered' from data through deep 
	architectures, and 
	(d) out system can trained in an end-to-end manner. 
	Building on standard tools from numerical analysis we develop very efficient algorithms for inference and learning,
	as well as a customized technique adapted to the semantic segmentation task. 
	This efficiency allows us to explore more sophisticated architectures for structured prediction in deep learning: 
	we introduce multi-resolution architectures to couple information across scales in a joint optimization framework, yielding systematic improvements. 
	We demonstrate the utility of our approach on the challenging VOC PASCAL 2012 image segmentation benchmark, 
	showing substantial improvements over strong baselines. We make all of our code and experiments available at \url{https://github.com/siddharthachandra/gcrf}.
\end{abstract}

\section{Introduction}
\label{sec:Introduction}
Over the last few years deep learning has resulted in dramatic progress in the task of semantic image segmentation. 
Early works on using CNNs as feature extractors \cite{lecun13,zoomout,hariharan} and combining them with standard superpixel-based front-ends gave substantial improvements 
over well-engineered approaches that used hand-crafted features. The currently mainstream approach is relying on `Fully' Convolutional Networks (FCNs) \cite{fcnn,farabet2012scene}, where CNNs are trained to provide fields of outputs used for pixelwise labeling.


\begin{figure*}[b!]
\centering
 \subfigure[Schematic of a fully convolutional neural network with a G-CRF module]{\includegraphics[width=0.95\textwidth]{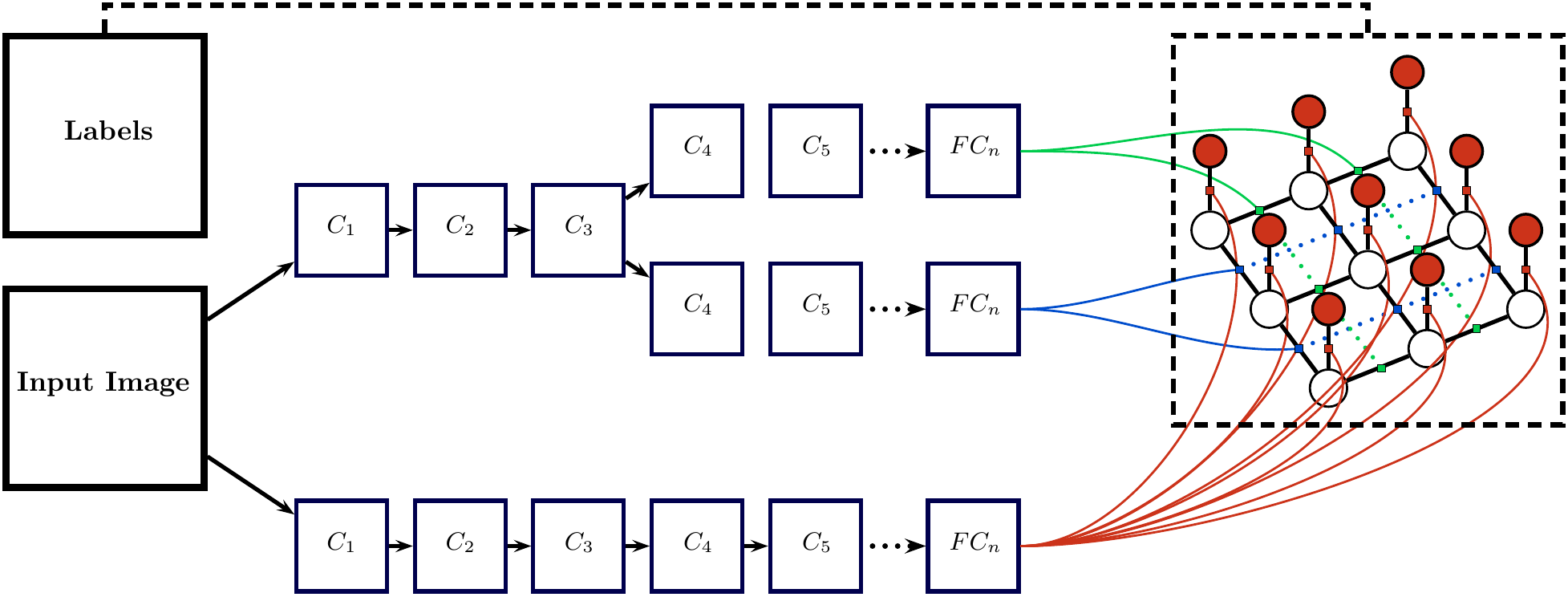}}\\
  \subfigure[\scriptsize{Input Image}]{\includegraphics[width=0.24\linewidth]{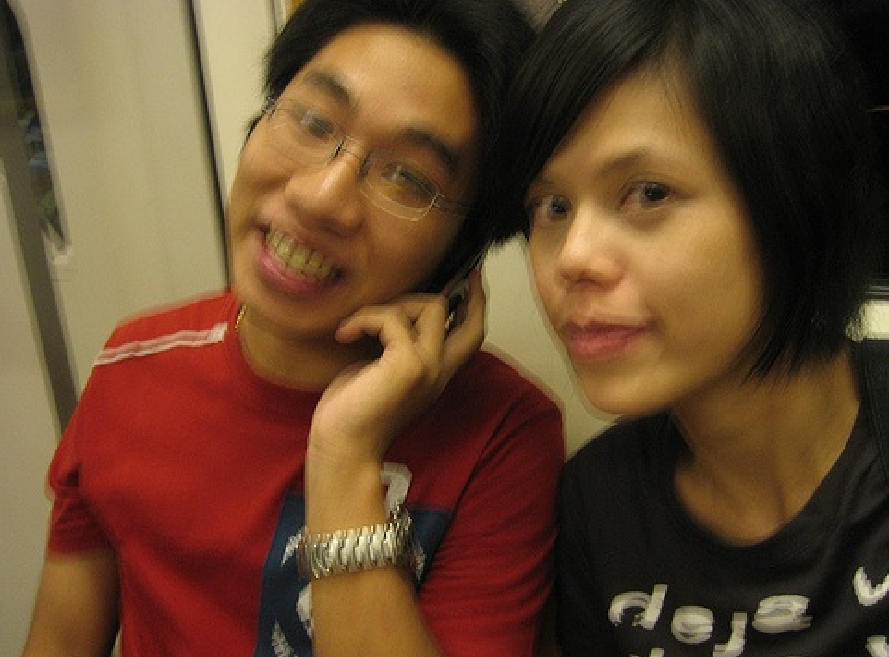}}
  \subfigure[\scriptsize{Person unary}]{\includegraphics[width=0.24\linewidth]{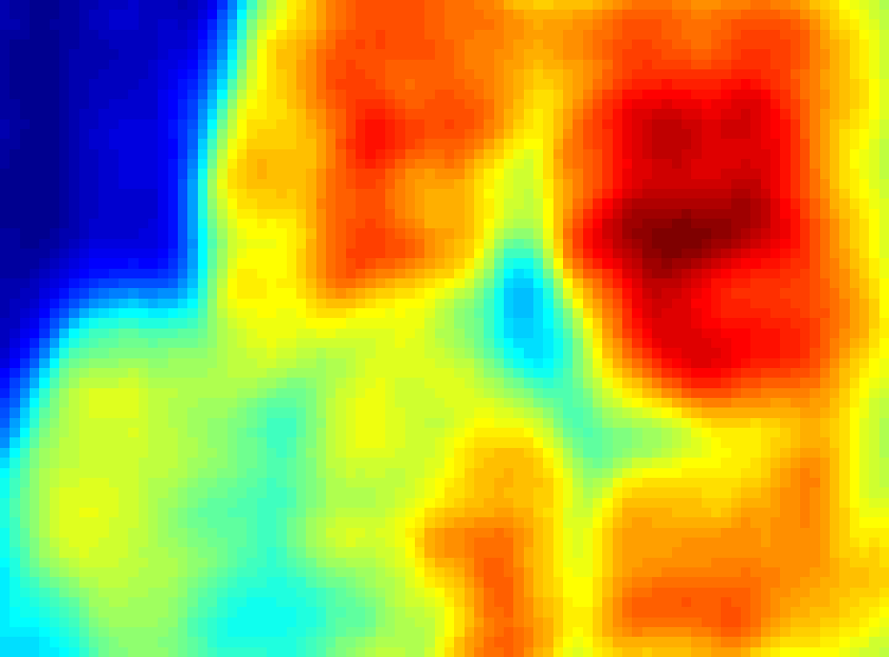}}
  \subfigure[\scriptsize{QO Output}]{\includegraphics[width=0.24\linewidth]{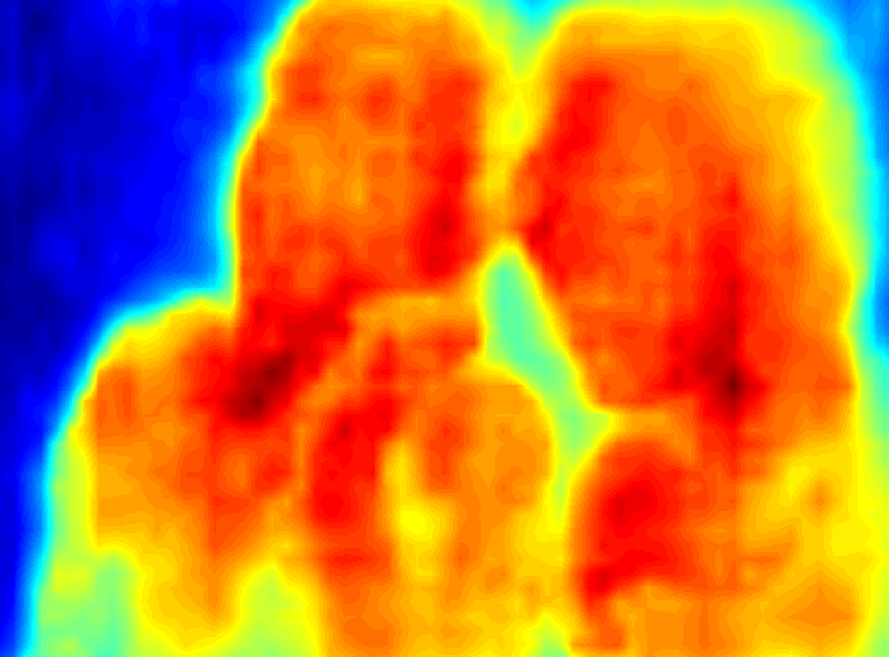}}
  \subfigure[\scriptsize{Person Probability}]{\includegraphics[width=0.24\linewidth]{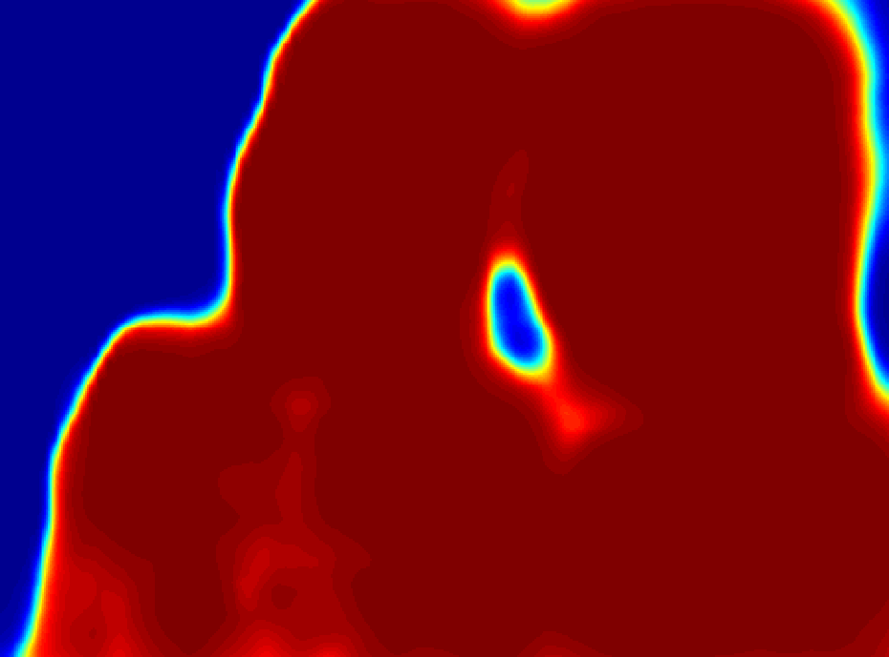}}
 \caption{\small{(a) shows a detailed schematic representation of our fully convolutional neural network with a G-CRF module.
 The G-CRF module is shown as the box outlined by dotted lines. The factor graph inside the G-CRF module shows a $4-$connected neighbourhood. 
 The white blobs represent pixels, red blobs represent unary factors, the green and blue squares represent vertical and horizontal connectivity factors.
  The input image is shown in (b). The network populates the unary terms (c), and horizontal and vertical pairwise
 terms.  The G-CRF module collects the unary and pairwise terms from the network and proposes an image hypothesis, i.e. scores (d) after inference. These scores 
 are finally converted to probabilities using the Softmax function (e), which are then thresholded to obtain the segmentation. It can
be seen that while the unary scores in (c) 
miss part of the torso because it is occluded behind the hand. 
The flow of information from the neighbouring region in the image, via the pairwise terms, encourages pixels in the occluded region to take the same label as the rest of the torso (d). Further it can be seen that the person boundaries are more pronounced in the output (d) due to pairwise constraints between pixels corresponding to the person and background classes.
 }}
 \label{fig:full}
\end{figure*}
A dominant research direction for improving semantic segmentation with deep learning is the combination of the powerful classification capabilities of FCNs with structured prediction  \cite{crfrnn,Vemulapalli_2016_CVPR,deeplab1,schwing,vemulapalli,IonescuVS15}, 
which aims at improving classification by capturing interactions between  predicted labels. 
 One of the first works in the direction of combining deep networks with structured prediction was \cite{deeplab1} which advocated the use of densely-connected conditional random fields (DenseCRF)~\cite{densecrf} to post-process an FCNN output so as to obtain a sharper segmentation the preserves image boundaries. This was then used by Zheng \etal~\cite{crfrnn} who combined DenseCRF with a CNN
into a single Recurrent Neural Network (RNN), accommodating the DenseCRF post processing in an end-to-end training procedure.

Most approaches for semantic segmentation perform structured prediction using approximate inference and learning \cite{schwing,couprie2012multi}.
For instance the techniques of \cite{crfrnn,Vemulapalli_2016_CVPR,deeplab1,vemulapalli} perform mean-field inference for a fixed number of 10 iterations.
Going for higher accuracy with more iterations 
could mean longer computation and eventually also memory bottlenecks:  back-propagation-through-time operates on  the intermediate `unrolled inference' results that have to be stored in  (limited) GPU memory. Furthermore, the non-convexity of the mean field objective means more iterations would only guarantee  convergence to a local minimum. 
The authors in \cite{Adelaide} use piecewise training with CNN-based pairwise potentials and three iterations of inference, while those in \cite{DPN} use highly-sophisticated modules, effectively learning to approximate mean-field inference. 
In  these two works a more pragmatic approach to inference is taken, considering it as a sequence of operations that need to be learned \cite{crfrnn}. 
These `inferning'-based approaches of combining learning and inference may be liberating, in the sense that one acknowledges and accommodates the approximations in the inference through end-to-end training. We show however here that exact inference and learning is feasible, while not making compromises in the model's expressive power. 



Motivated by \cite{TappenLAF07,rtf}, our starting point in this work is the observation that a particular type of graphical model, the Gaussian Conditional Random Field (G-CRF), 
allows us to perform exact and efficient Maximum-A-Posteriori (MAP) inference. 
Even though Gaussian Random Fields are unimodal and as such less expressive, Gaussian \emph{Conditional} Random Fields are unimodal \emph{conditioned on the data}, effectively
reflecting the fact that given the image one solution dominates the posterior distribution.
The G-CRF model thus allows us to construct rich expressive structured prediction models that still lend
themselves to efficient inference. In particular, the log-likelihood of the G-CRF posterior has the form of a quadratic energy function which captures unary and pairwise interactions between random
variables. There are two advantages to using a quadratic function: (a) unlike the energy of general graphical models, a quadratic  function has a unique global minimum if the system matrix is
positive definite, and (b) this unique minimum can be efficiently found by solving a system of linear equations.
We can actually discard the probabilistic underpinning of the G-CRF and understand G-CRF inference as an energy-based model, casting structured prediction as quadratic optimization (QO). 

G-CRFs were exploited for instance in the regression tree fields model of Jancsary \etal~\cite{rtf}  where decision trees were used to construct G-CRF's and address a host of  vision tasks, including inpainting, segmentation and pose estimation.
In independent work \cite{Vemulapalli_2016_CVPR} proposed a similar approach for the task of image segmentation with CNNs, where 
as  in \cite{Adelaide,DPN,vu2015context} FCNs are augmented with discriminatively trained convolutional layers that model and enforce pairwise consistencies between neighbouring regions.

One major difference to \cite{Vemulapalli_2016_CVPR}, as well as other prior works~\cite{crfrnn,deeplab1,vemulapalli,Adelaide,DPN}, is that
we use exact inference and do not use back-propagation-through-time
during training. In particular building on the insights of \cite{TappenLAF07,rtf}, we observe that the MAP solution, as well as  the gradient of our objective with respect to the inputs of our structured prediction module can be obtained through the solution of linear systems.
Casting the learning and inference tasks in terms of linear systems allows us to exploit the wealth of tools from numerical analysis. As we show in Sec.~ \ref{sec:optimization}, for Gaussian CRFs sequential/parallel mean-field inference amounts to solving a linear system using the classic Gauss-Seidel/Jacobi algorithms respectively. Instead of these under-performing methods we use conjugate gradients which allow us to perform exact inference and back-propagation in a small 
number (typically 10) iterations, with a negligible cost ($0.02$s for the general case in Sec.~\ref{section:formulation}, and $0.003$s for the simplified formulation in Sec.~\ref{sec:simpleqo}) when implemented on the GPU. 


Secondly, building further on the connection between MAP inference and linear system solutions, we  propose memory- and time-efficient algorithms for weight-sharing (Sec.~\ref{sec:simpleqo}) and multi-scale inference
 (Sec.~\ref{sec:multires}).  In particular, in Section \ref{sec:simpleqo} we show that one can further reduce the memory footprint and computation demands of our method by introducing a Potts-type structure in the pairwise term. This results in multifold accelerations, while delivering  results that are competitive to the ones obtained with the unconstrained pairwise term. 
In Sec.~\ref{sec:multires} we show that our approach allows us to work with arbitrary neighbourhoods that go beyond 
the common $4-$connected neighbourhoods. In particular we  explore the merit of using multi-scale networks, 
  where variables computed from different image scales interact with each other. This gives rise to a flow of information across different-sized neighborhoods. We show experimentally that this yields substantially 
  improved results over single-scale baselines.

In \refsec{section:formulation} we describe
our approach in detail, and derive the expressions for weight update rules for parameter learning that are
used to train our networks in an end-to-end manner.
In  \refsec{sec:optimization} we analyze the efficiency of the linear system solvers and present our multi-resolution structured prediction algorithm. 
In \refsec{sec:experiments} we report consistent improvements over well-known baselines and state-of-the-art results on the VOC PASCAL  test set.

\newcommand{\refeq}[1]{Eq.~\ref{#1}}
\section{Quadratic Optimization Formulation}
\label{section:formulation}


We now describe our approach. 
Consider an image $\mathcal{I}$ containing $P$ pixels. Each pixel $p \in \{p_1,\ldots,p_P\}$ can take a label $l \in \{1,\ldots,L\}$. 
Although our objective is to assign discrete labels to the pixels, we phrase our problem as a continuous inference task. Rather than performing a discrete inference task that delivers one label per variable, we use
a continuous function of the form ${\bf x}(p,l)$ which gives a score for each pairing of a pixel to a label. 
This score can be intuitively
understood as being proportional to the log-odds for the pixel $p$ taking the label $l$, if a `softmax' unit is used to post-process ${\bf x}$.

We denote the pixel-level ground-truth labeling by a discrete valued vector $\mathbf{y} \in \mathbb{Y}^{P}$ where $\mathbb{Y} \in \{1,\ldots,L\}$, and the inferred hypothesis by
a real valued vector $\mathbf{x} \in \mathbb{R}^{N}$, where $N = P\times L$. 
Our formulation is posed as an energy minimization problem. In the following subsections, we
describe the form of the energy function, the inference procedure, and the parameter learning approach, followed by some technical
details pertinent to using our framework in a fully convolutional neural network. Finally, we describe a simpler formulation with pairwise weight sharing  which 
achieves competitive performance while being substantially faster. Even though our inspiration was from the probabilistic approach to structured prediction (G-CRF),
from now on we treat our structured prediction technique as a Quadratic Optimization (QO) module, and will refer to it as QO henceforth.

\subsection{Energy of a hypothesis} We define the energy of a hypothesis in terms of a function of the following form:

\begin{equation}
 E({\bf x}) = \frac{1}{2} {\bf x}^T (A+ \lambda \textbf{I}) {\bf x} - B{\bf x}
 \label{eqn:energy}
\end{equation}

\noindent where $A$ denotes the symmetric $N\times N$ matrix of pairwise terms, and $B$ denotes the $N\times 1$ vector of unary terms.
In our case, as shown in Fig.~\ref{fig:full}, the pairwise terms $A$ and the unary terms $B$ are learned from the data using a fully convolutional network. In particular and as illustrated in 
Fig.~\ref{fig:full},
$A$ and $B$ are the outputs of the pairwise and unary streams of our network, computed by a forward pass on the input image. These unary 
and pairwise terms are then combined by the QO module to give the final per-class scores for each pixel in the image. As we show below, during training we can easily obtain the gradients of the output with respect to the $A$ and $B$ terms, allowing us to train the whole network end-to-end. 

Eq. \ref{eqn:energy} is a standard way of expressing the energy of a system with unary and pair-wise interactions among the random variables \cite{rtf} in a vector labeling task. 
We chose this function primarily because it has a unique global minimum and allows for exact inference, alleviating the need for approximate inference. 
Note that in order to make the matrix $A$ strictly positive definite, we add to it $\lambda$ times the Identity Matrix ${\textbf I}$, where $\lambda$ is a design parameter set empirically in the experiments.

\subsection{Inference} Given $A$ and $B$, inference involves solving
for the value of ${\bf x}$ that minimizes the energy function in Eq.~\ref{eqn:energy}.
 If ($A+\lambda {\textbf I}$) is symmetric positive definite, then $E({\bf x})$ has a unique global minimum \cite{conjugategradient} at:
\begin{equation}
 ( A + \lambda \mathbf{I} ) \mathbf{x} = B\text{.}
 \label{eqn:linearSolver}
\end{equation}
As such, inference  is exact and efficient, only involving a system of linear equations.

\subsection{Learning A and B}
\label{subsec:learningAB}
Our model parameters $A$ and $B$ are learned in an end-to-end fashion via the back-propagation method.
In the back-propagation training paradigm each module or \emph{layer} in the network receives
the derivative of the final loss $\mathcal{L}$ with respect to its output $\textbf{x}$, denoted by
$\frac{\partial \mathcal{L}}{\partial \textbf{x}}$, from the layer above. $\frac{\partial \mathcal{L}}{\partial \textbf{x}}$ is also referred to as the
gradient of $\textbf{x}$. The module then computes the gradients of its inputs and propagates them down through the
network to the layer below.

To learn the parameters $A$ and $B$ via back-propagation, we require the expressions of gradients of $A$ and $B$, i.e. $\frac{\partial \mathcal{L}}{\partial A}$ and $\frac{\partial \mathcal{L}}{\partial B}$ respectively. We now derive these expressions.

\subsubsection{Derivative of Loss with respect to B}
\label{subsubsection:derivative_b}
To compute the derivative of the loss with respect to B, we use the chain rule of differentiation: $\frac{\partial \mathcal{L}}{\partial \mathbf{x}}  = \frac{\partial \mathcal{L}}{\partial B}\frac{\partial B}{\partial \mathbf{x}}  $.
Application of the chain rule yields the following closed form expression, which is a system of linear equations:
%
%
\begin{equation}
 ( A + \lambda \mathbf{I} ) \frac{\partial \mathcal{L}}{\partial B} = \frac{\partial \mathcal{L}}{\partial \textbf{x}}\text{.}
 \label{eqn:dldbres}
\end{equation}
When training a deep network, the right hand side $\frac{\partial \mathcal{L}}{\partial B}$ is delivered by the layer above, and the derivative on the left hand side is sent to the unary layer below. 

\subsubsection{Derivative of Loss with respect to A}
The expression for the gradient of $A$ is derived by using the chain rule of differentiation again: $\frac{\partial \mathcal{L}}{\partial A} =\frac{\partial \mathcal{L}}{\partial \mathbf{x}}  \frac{\partial \mathbf{x}}{\partial A}$. 

Using the expression $ \frac{\partial \mathbf{x}}{\partial A} = \frac{\partial}{\partial A} (A + \lambda \mathbf{I} )^{-1} B $, substituting 
 $  \frac{\partial}{\partial A} (A + \lambda \mathbf{I} )^{-1} = - (A + \lambda \mathbf{I} )^{-T} \otimes (A + \lambda \mathbf{I} )^{-1}$,
 and simplifying the right hand side,
 we arrive at the following expression:
 
\begin{equation}
 \frac{\partial \mathcal{L}}{\partial A} = - \frac{\partial \mathcal{L}}{\partial B}  \otimes \mathbf{x}\text{,}
 \label{eqn:dlda2_1}
\end{equation}
where $\otimes$ denotes the kronecker product.
Thus, the gradient of $A$ is given by the negative of the kronecker product of the output $\textbf{x}$ and the gradient of $B$.

%
%
\subsection{Softmax Cross-Entropy Loss}
Please note that while in this work we use the QO module as the penultimate layer of the network, followed by the
softmax cross-entropy loss, it can be used at any stage in a network and not only as the final classifier.
We now give the expressions for the softmax cross-entropy loss and its derivative for sake of completeness.

The image hypothesis is a scoring function of the form $\mathbf{x}(p,l)$. For brevity, we denote the hypothesis concerning a single pixel by $\mathbf{x}(l)$.
The softmax probabilities for the labels are then given by 
$p_l = \frac{e^{\mathbf{x}(l)}}{\sum_L e^{\mathbf{x}(l)}}\text{.}$
These probabilities are penalized by the cross-entropy loss
defined as
$\mathcal{L} = -\sum_l \mathbf{y}_l \log p_l$, 
where $\mathbf{y}_l$ is the ground truth indicator function for the ground truth label $l^*$, i.e. $\mathbf{y}_l = 0$ if $l \neq l^*$, and $\mathbf{y}_l = 1$ otherwise.
Finally the derivative of the softmax-loss with respect to the input is given by:
$\frac{\partial \mathcal{L}}{\partial \mathbf{x}(l)} = p_l - y_l$.

\subsection{Quadratic Optimization with Shared Pairwise Terms}
\label{sec:simpleqo}
We now describe a simplified \emph{QO} formulation with shared pairwise terms which is significantly faster in practice than the one described above.
We denote by $A_{p_i,p_j}(l_i,l_j)$ the pairwise energy term for pixel $p_i$ taking the label $l_i$, and pixel $p_j$ taking the label $l_j$.
In this section, we propose a \emph{Potts}-type pairwise model, described by the following equation:
\begin{equation}
 A_{p_i,p_j}(l_i,l_j) = \left\{
\begin{array}{ll}
      0 & l_i = l_j \\
      A_{p_i,p_j} & l_i \neq l_j{.} \\
\end{array} 
\right\}
\end{equation}

In simpler terms, unlike in the general setting, the pairwise terms here depend on whether the pixels take the same label or not, and not on the particular labels they take.
Thus, the pairwise terms are \emph{shared} by different pairs of classes. 
While in the general setting we learn $PL \times PL$ pairwise terms, here we learn only $P\times P$ terms. To derive the inference and gradient equations after this simplification,
we rewrite our inference equation $\left(A+\lambda \textbf{I}\right) \textbf{x} = B$ as,

\begin{equation}
\begin{bmatrix} \lambda \textbf{I} & \hat{A} & \cdots & \hat{A} \\ \hat{A} & \lambda \textbf{I} & \cdots & \hat{A} \\ & & \vdots & \\ \hat{A} & \hat{A} & \cdots & \lambda \textbf{I} \end{bmatrix} \times 
\left[ \begin{array}{c} {\textbf{x}_1} \\ \textbf{x}_2 \\ \vdots \\ \textbf{x}_L \end{array} \right] =  
\left[ \begin{array}{c} \textbf{b}_1 \\ \textbf{b}_2 \\ \vdots \\ \textbf{b}_L \end{array} \right]
\label{eq:matrixform}
\end{equation}
where $\textbf{x}_k$, denotes the vector of scores for all the pixels for the class $k \in \{1,\cdots,L\}$. The per-class unaries are denoted by $\textbf{b}_k$, and the pairwise terms $\hat{A}$ are 
shared between each pair of classes. The equations that follow are derived by specializing the general inference (Eq.~\ref{eqn:linearSolver}) and gradient equations (Eq.~\ref{eqn:dldbres},\ref{eqn:dlda2_1}) to this particular setting.
Following simple manipulations, the inference procedure becomes a two step process where we first compute the sum of our 
scores $\sum_i \textbf{x}_i$, followed by $\textbf{x}_k$, i.e. the scores for the class $k$ as:

\noindent\begin{tabularx}{\textwidth}{@{}XX@{}}
\begin{equation}
\left(\lambda \textbf{I} + \left(L - 1\right)\hat{A}\right) \sum_{i} \textbf{x}_i = \sum_i \textbf{b}_i \text{,}
\label{eq:sumxk}
\end{equation} &
\begin{equation}
(\lambda \textbf{I}-\hat{A})\textbf{x}_k = \textbf{b}_k - \hat{A}\sum_i \textbf{x}_i \text{.}
\label{eq:xk}
\end{equation} 
\end{tabularx}

Derivatives of the unary terms with respect to the loss are obtained by solving:

\noindent\begin{tabularx}{\textwidth}{@{}XX@{}}
\begin{equation}
\left(\lambda \textbf{I} + \left(L - 1\right)\hat{A}\right) \sum_{i} \frac{\partial \mathcal{L}}{\partial \textbf{b}_i} = \sum_i \frac{\partial \mathcal{L}}{\partial \textbf{x}_i} \text{,}
\label{eq:sumdiffeqn}
\end{equation} &
\begin{equation}
(\lambda \textbf{I}-\hat{A})\frac{\partial \mathcal{L}}{\partial \textbf{b}_k} = \frac{\partial \mathcal{L}}{\partial \textbf{x}_k} - \hat{A} \sum_{i} \frac{\partial \mathcal{L}}{\partial \textbf{b}_i} \text{.}
\label{eq:diffbk}
\end{equation} 
\end{tabularx}

Finally, the gradients of $\hat{A}$ are computed as 
\begin{equation}
\frac{\partial \mathcal{L}}{\partial \hat{A}} = \sum_{k} \frac{\partial \mathcal{L}}{\partial \textbf{b}_k} \otimes \sum_{i\neq k} \textbf{x}_i \text{.}
\end{equation}
Thus, rather than solving a system with $A \in \mathbb{R}^{PL \times PL}$, we solve $L+1$ systems with $\hat{A} \in \mathbb{R}^{P\times P}$. In our case, where $L = 21$ for $20$ object classes and $1$ background class, 
this simplification empirically reduces the inference time by a factor of $6$, and the overall training time by a factor of $3$.
We expect even larger acceleration for the MS-COCO dataset which has $80$ semantic classes. 
Despite this simplification, the results are competitive to the general setting as shown in Sec.~\ref{sec:experiments}.

\section{Linear Systems for Efficient and Effective Structured Prediction}
\label{sec:optimization}
Having identified that both the inference problem in \refeq{eqn:linearSolver} and computation of pairwise gradients in \refeq{eqn:dldbres} require the
solution of a linear system of equations, we 
now discuss  methods for accelerated inference that rely on standard numerical analysis techniques for linear systems  \cite{PressTVF92,Golub96}. 
Our  main contributions consist in (a) using fast linear system solvers that exhibit 
 fast convergence (\refsec{sec:solvers})  and (b) performing inference on multi-scale graphs by constructing block-structured linear systems (\refsec{sec:multires}).
 
 Our contributions in (a) indicate that standard conjugate gradient based linear system solvers can be up to 2.5 faster than the solutions one could get by a 
 naive application of parallel mean-field when implemented on the GPU. Our contribution in (b)  aims at accuracy rather than efficiency, and is experimentally validated in \refsec{sec:experiments}

\subsection{Fast Linear System Solvers}
\label{sec:solvers}

\begin{figure}[]
\centering

\begin{minipage}{0.4\textwidth} 
 
\centering
\begin{minipage}{\linewidth} 
\centering
    \begin{tabular}{l|r }
    \hline
    Method & Iterations \\ \hline 
    Jacobi & 24.8 \\ \hline
    Gauss Siedel & 16.4 \\ \hline
    GMRES & 14.8 \\ \hline
    Conjugate Gradient & 13.2 \\ \hline
    \end{tabular}\\[1mm](a) Linear Solver Statistics
    \label{table:solvers}
  \end{minipage} 
  \\[2mm]
\end{minipage} 
\begin{minipage}{0.5\textwidth}
\centering
\includegraphics[width=0.8\textwidth]{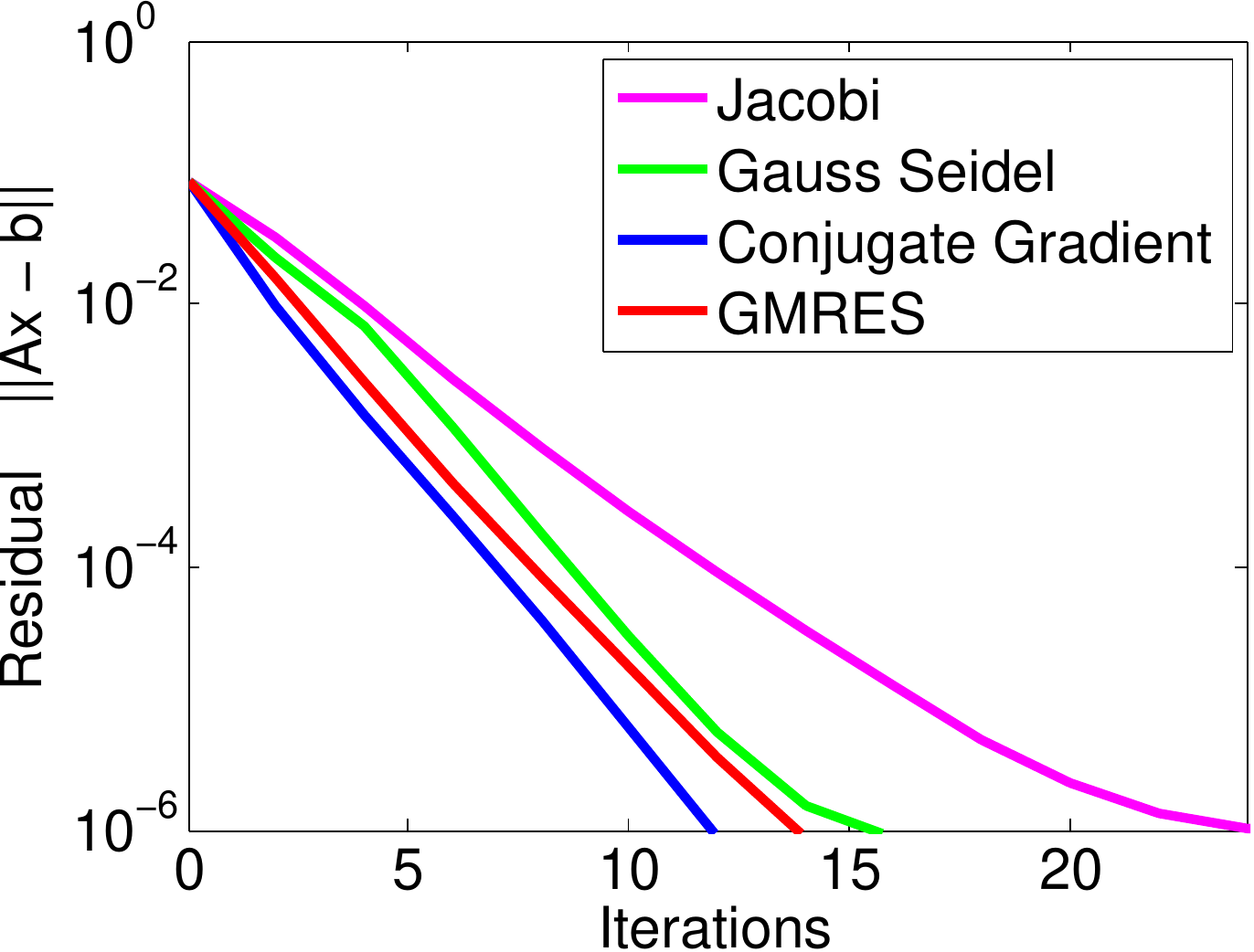}
\\(c) Iterative Solvers Convergence
\end{minipage} 
\caption{The table in (a) shows the average number of iterations required by various algorithms,
namely Jacobi, Gauss Seidel, Conjugate Gradient, and Generalized Minimal Residual (GMRES) iterative methods
to converge to a residual of tolerance $10^{-6}$. 
Figure (b) shows a plot demonstrating the convergence of these iterative solvers.
The conjugate gradient method outperforms the other competitors in terms of number of iterations taken to converge.
\label{fig:plot_solversa}
}
\end{figure}
		
%
%
%
%

\mycomment{
The linear system solver algorithm that we employ will directly affect the computational efficiency of the quadratic optimization formulation,
and impact its applicability to general problems, and its use in networks for a variety of tasks.
}

The computational cost of solving the linear system of equations in \refeq{eqn:linearSolver} and \refeq{eqn:dldbres} depends on the size of the matrix $A$, i.e. $N\times N$,
and its sparsity pattern. In our experiments, while $N \sim 10^5$, the matrix $A$ is quite sparse, 
since we deal with small $4-$connected, $8-$connected and $12-$connected neighbourhoods.
While a number of direct linear system solver methods exist, the sheer size of the system matrix $A$ renders
them prohibitive, because of large memory requirements. For large problems, a number of iterative methods exist, which
require less memory, come with convergence (to a certain tolerance) guarantees under certain conditions, and can be faster than direct methods.
In this work, we considered the \emph{Jacobi, Gauss-Seidel, Conjugate Gradient, and Generalized Minimal Residual} (GMRES) methods  \cite{PressTVF92}, as candidates
for iterative solvers. The table in Fig. \ref{fig:plot_solversa} (a) shows the average number of iterations required by the aforementioned methods for solving the
inference problem in Eq.~\ref{eqn:linearSolver}. We used $25$ images in this analysis, and a tolerance of $10^{-6}$. 
Fig.~\ref{fig:plot_solversa} shows
the convergence of these methods for one of these images. 
Conjugate gradients clearly stand out as being the fastest of these methods, so our following results use the conjugate gradient method. 
Our findings are consistent with those of Grady in \cite{grady}.

As we show below, mean-field inference for the Gaussian CRF  can be understood as solving the linear system of \refeq{eqn:linearSolver}, namely
parallel mean-field amounts to using the Jacobi algorithm while sequential mean-field amounts to 
 using the Gauss-Seidel algorithm, which are the two weakest baselines in our comparisons. 
This indicates that by resorting to tools for solving linear systems
we have introduced faster alternatives to those suggested by mean field.

In particular the \emph{Jacobi} and \emph{Gauss-Seidel} methods solve a system of linear equations $A\textbf{x} = B$
by generating a sequence of approximate solutions $\left\lbrace \textbf{x}^{(k)} \right\rbrace$,
where the current solution $\textbf{x}^{(k)}$ determines the next solution $\textbf{x}^{(k+1)}$.

The update equation for the \emph{Jacobi} method \cite{matrixcomputation} is given by
\begin{equation}
x^{(k+1)}_i \gets  \frac{1}{a_{ii}} \left\lbrace b_i - \sum_{j \neq i} a_{ij}x^{(k)}_j \right\rbrace \text{.}
\label{eqn:jacobi}
\end{equation}

The updates in Eq.~\ref{eqn:jacobi} only use the previous solution $\textbf{x}^{(k)}$, ignoring the most recently
available information. For instance, $x_1^{(k)}$ is used in the calculation of $x_2^{(k+1)}$, even though $x_1^{(k+1)}$
is known. This allows for parallel updates for $\textbf{x}$. In contrast, the \emph{Gauss-Seidel} \cite{matrixcomputation} method always uses the most
current estimate of $x_i$ as given by:
\begin{equation}
x^{(k+1)}_i \gets \frac{1}{a_{ii}} \left\lbrace b_i - \sum_{j < i} a_{ij}x^{(k+1)}_j - \sum_{j > i} a_{ij}x^{(k)}_j\right\rbrace \text{.}
\label{eqn:gauss}
\end{equation}

As in \cite{GMRFbook},
the Gaussian Markov Random Field (GMRF) in its canonical form is expressed as
$\pi(\textbf{x}) \propto \text{exp}\left\lbrace \frac{1}{2} \textbf{x}^T \Theta \textbf{x} + \theta^T\textbf{x} \right\rbrace$, where
$\theta$ and $\Theta$ are called the canonical parameters associated with the multivariate Gaussian distribution $\pi(\textbf{x})$.
The update equation corresponding to mean-field inference is given by~\cite{jordan},
\begin{equation}
\mu_i \gets - \frac{1}{\Theta_{ii}} \left\lbrace \theta_i + \sum_{j \neq i} \Theta_{ij}\mu_j \right\rbrace \text{,}
\label{eqn:meanfield}
\end{equation}
The expression in Eq.~\ref{eqn:meanfield} is exactly the expression for the \emph{Jacobi} iteration (Eq.~\ref{eqn:jacobi}), or the \emph{Gauss-Seidel} iteration in Eq.~\ref{eqn:gauss} for solving the linear system  $\mu = -\Theta^{-1}\theta$,
depending on whether we use sequential or parallel updates.

One can thus understand sequential and parallel mean-field inference and learning algorithms
as relying on weaker system solvers than the conjugate gradient-based ones we propose here. The connection is accurate for Gaussian CRFs, as in our work and \cite{Vemulapalli_2016_CVPR}, and only intuitive for Discrete CRFs used in \cite{crfrnn,deeplab1}. 

\subsection{Multiresolution graph architecture}
\label{sec:multires}
\begin{figure}
		\centering
		\includegraphics[width=0.8\textwidth]{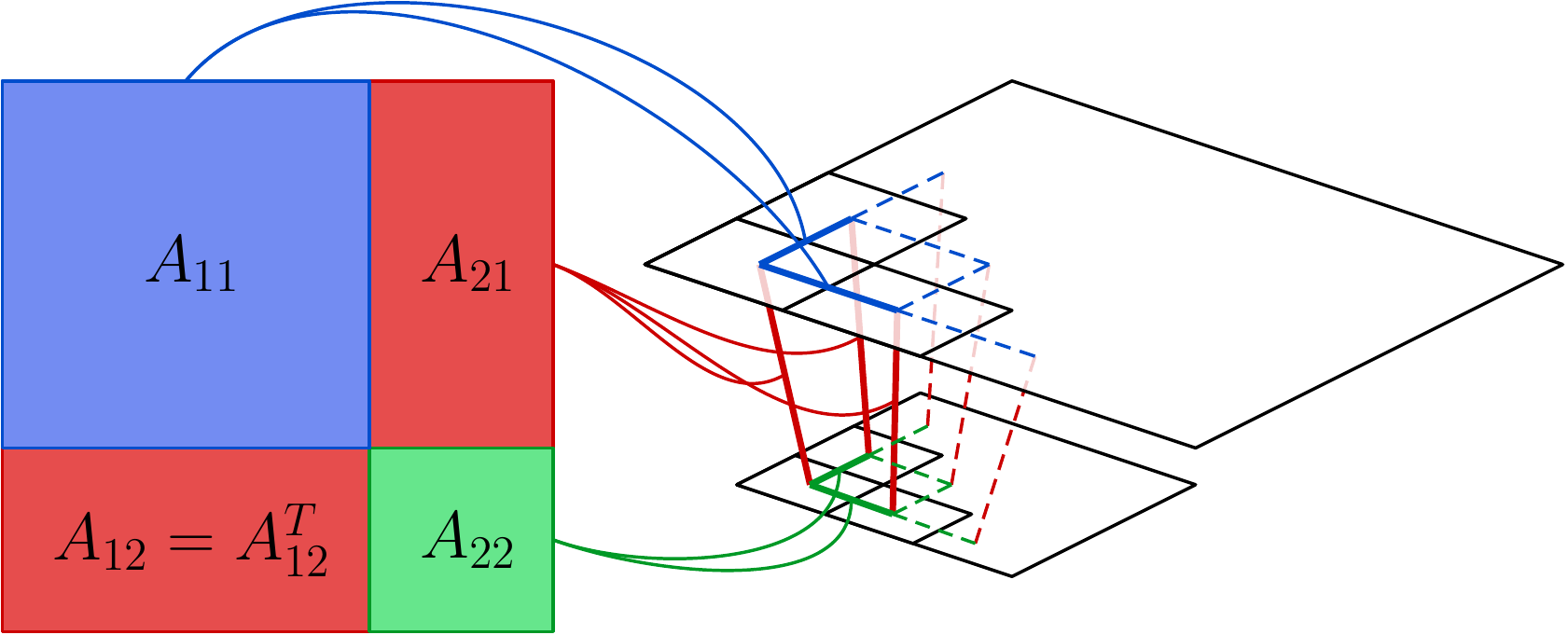}
		\caption{Schematic diagram of matrix $A$ for the multi-resolution formulation in \refsec{sec:multires}. In this example, we have the input image at $2$ resolutions. 
		The pairwise matrix $A$ contains two kinds of pairwise interactions: (a) neighbourhood interactions between pixels at the same resolution (these interactions are shown as the blue and green squares), 
		and (b) interactions between the same image region at two resolutions (these interactions are shown as red rectangles). While interactions of type (a) encourage the pixels in a neighbourhood to take
		the same or different label, the interactions of type (b) encourage the same image region to take the same labels at different resolutions.
		}
		\label{fig:plot_solversb}
\end{figure}

We now turn to incorporating computation from multiple scales in a single system. Even though CNNs are designed to be largely scale-invariant, it has been repeatedly reported \cite{ChenYWXY15,edg} that fusing information from a CNN operating at multiple scales can improve image labeling performance.
These results have been obtained for feedforward CNNs - we consider how these could be extended to CNNs with lateral connections, as in our case. 
A simple way of achieving this would be to use multiple image resolutions, construct one structured prediction module per resolution, train these as disjoint networks, and average the final results. 
This  amounts to solving three decoupled systems which by itself yields a certain improvement as reported in \refsec{sec:experiments} 

We advocate however  a richer connectivity that couples the scale-specific systems, allowing information to flow across scales. As illustrated in \reffig{fig:plot_solversb} the resulting linear system captures the following multi-resolution interactions simultaneously: (a) pairwise constraints between pixels at each resolution, and (b) pairwise
constraints between the same image region at two different resolutions. These inter-resolution pairwise terms connect a pixel in the image at one resolution, to the pixel it would spatially correspond to at another resolution. The inter-resolution connections
help enforce a different kind of pairwise consistency: rather than encouraging pixels in a neighbourhood to have the same/different label, these encourage image regions to have the same/different labels across resolutions. This is experimentally validated in \refsec{sec:experiments} to outperform the simpler multi-resolution architecture outlined above. 


\subsection{Implementation Details and Computational Efficiency}
Our implementation is fully GPU based, and implemented using the \emph{Caffe} library. Our network processes input images of size $865\times 673$, and delivers results at a resolution that is $8$ times smaller, as in \cite{deeplab1}.
The input to our QO modules is thus a feature map of size $109 \times 85$. While the testing time per image for our methods is between $0.4-0.7$s per image,
our inference procedure only takes $\sim 0.02$s for the general setting in Sec.~\ref{section:formulation}, and 
$0.003$s for the simplified formulation (Sec.~\ref{sec:simpleqo}). This is significantly faster than 
dense CRF postprocessing, which takes $2.4$s for a $375 \times 500$ image on a CPU and the $0.24$s on a GPU.
Our implementation uses the highly optimized \emph{cuBlas} and \emph{cuSparse} libraries for linear algebra on large sparse
matrices. The \emph{cuSparse} library requires the matrices to be in the compressed-storage-row (CSR) format in order to fully optimize linear
algebra for sparse matrices. Our implementation caches the indices of the CSR matrices, and as such their computation time is not taken into account 
in the calculations above, since their computation time is zero for streaming applications, or if the images get warped to a canonical size. 
In applications where images may be coming at different dimensions, considering that the
 indexes have been precomputed for the changing dimensions, an additional overhead of $\sim 0.1$s per image is incurred to read the binary files containing the cached 
 indexes from the hard disk (using an SSD drive could further reduce this). Our code and experiments are publicly available at \url{https://github.com/siddharthachandra/gcrf}.

\section{Experiments}
\label{sec:experiments}
In this section, we describe our experimental setup, network architecture and results.

\textbf{Dataset.} We evaluate our methods on the \emph{VOC PASCAL 2012 image segmentation benchmark}. This benchmark uses the VOC PASCAL 2012 dataset, which consists of
$1464$ training and $1449$ validation images with manually annotated pixel-level labels for $20$ foreground object classes, and $1$ background class. 
In addition, we exploit the additional
pixel-level annotations provided by \cite{hariharan}, obtaining $10582$ training images in total. The test set has $1456$ unannotated images. The evaluation
criterion is the pixel intersection-over-union (IOU) metric, averaged across the $21$ classes.

\textbf{Baseline network (basenet).} Our basenet is based on the Deeplab-LargeFOV network from \cite{deeplab1}. As in \cite{edg}, we extend it to get a multi-resolution network,
which operates at three resolutions with tied weights. More precisely, our network downsamples the input image by factors of $2$ and $3$
and later \emph{fuses} the downsampled activations with the original resolution via concatenation followed by convolution. The layers at three resolutions share weights. This acts like a strong baseline for a purely feedforward network.
Our basenet has $49$ convolutional layers, $20$ pooling layers, and was pretrained on the MS-COCO 2014 trainval dataset~\cite{coco}. The initial learning rate was set to $0.01$ and decreased by a factor of $10$ at $5$K iterations. It was trained for $10$K iterations.

\textbf{QO network.} We extend our basenet to accommodate the binary stream of our network. Fig.~\ref{fig:full} shows a rough schematic diagram of our network. 
The basenet forms the unary stream of our QO network, while the pairwise stream is composed by concatenating the $3^{rd}$ pooling layers of the three resolutions followed by \emph{batch normalization} and two convolutional layers. Thus, in Fig.~\ref{fig:full}, layers $C_1-C_3$ are shared by the unary and pairwise streams in our experiments. Like our basenet, the QO networks were trained for $10$K iterations; The initial learning rate was set to $0.01$ which was decreased by a factor of $10$ at $5$K iterations.
We consider three main types of QO networks: plain ($QO$), shared weights ($QO^{s}$) and multi-resolution ($QO^{mres}$).


\subsection{Experiments on train+aug - val data} 
\label{subsec:prelim}
In this set of experiments we train our methods on the \emph{train+aug} images, and evaluate them on the \emph{val} images.
All our images were upscaled to an input resolution of $865 \times 673$. 
The hyper-parameter $\lambda$ was set to $10$ to ensure positive definiteness.
We first study the effect of having larger neighbourhoods among image regions, thus allowing richer connectivity. More precisely, we study three kinds of connectivities: (a) $4-$connected (QO$_4$), where each pixel
is connected to its left, right, top, and bottom neighbours, (b) $8-$connected (QO$_8$), where each pixel is additionally connected to the $4$ diagonally adjacent neighbours, and (c) $12-$connected (QO$_{12}$), where each pixel
is connected to $2$ left, right, top, bottom neighbours besides the diagonally adjacent ones. Table \ref{table:richer} demonstrates that while there are improvements in performance upon increasing
connectivities, these are not substantial. Given that we obtain diminishing returns, rather than trying even larger neighbourhoods to improve performance, we focus on increasing the richness of the representation 
by incorporating information from various scales. As described in \refsec{sec:multires}, there are two ways to incorporate information from multiple scales; the simplest is to have one QO unit per resolution ($QO^{res}$), 
thereby enforcing pairwise consistencies individually at each resolution before fusing them, 
while the more sophisticated one is to have information flow both within and across scales, amounting to a joint multi-scale CRF inference task, illustrated in \reffig{fig:plot_solversb}. 
In Table \ref{table:qon}, we compare $4$ variants of our QO network: (a) QO (Sec.~\ref{section:formulation}), (b) QO with shared weights (Sec.~\ref{sec:simpleqo}),
(c) three QO units, one per image resolution, and (d) multi-resolution QO (Sec.~\ref{sec:multires}). It can be seen that our weight sharing simplification, while being significantly faster, also gives better results than QO. 
Finally, the multi-resolution framework outperforms the other variants, indicating that having information flow both within and across scales is desirable, and a unified multi-resolution framework is better than merely averaging QO scores from different
image resolutions.

\begin{table}[tb]
\begin{minipage}{0.3\linewidth}
\centering
\begin{tabular}{l|r|r|r }
\hline
Method & QO$_4$ & QO$_8$ & QO$_{12}$ \\ \hline
IoU & 76.36 & 76.40 & 76.42 \\ \hline
\end{tabular}
\vspace{2mm}
\caption{Connectivity}
\label{table:richer}
\end{minipage}
\begin{minipage}{0.7\linewidth}
\centering
\begin{tabular}{l|r|r|r|r}
\hline
Method & QO & QO$^{s}$ & QO$^{res}$ & QO$^{mres}$ \\ \hline
IoU & 76.36 & 76.59 & 76.69 &  76.93 \\ \hline
\end{tabular}
\vspace{2mm}
\caption{Comparison of $4$ variants of our G-CRF network.}
\label{table:qon}
\end{minipage}
\end{table}
%
\mycomment{
Finally, a more sophisticated approach is to have a multi-resolution QO (QO$^{mres}$). QO$^{mres}$ captures the following multi-resolution interactions simultaneously: (a) pairwise constraints between pixels at each resolution, and (b) pairwise
constraints between the same image region at two different resolutions. These inter-resolution pairwise terms connect a pixel in the image at one resolution, to the pixel it would spatially correspond to at another resolution. The inter-resolution connections
help enforce a different kind of pairwise consistency: rather than encouraging pixels in a neighbourhood to have the same/different label, these encourage image regions to have the same/different labels across resolutions. Table \ref{table:qon}
reports the performance of the different kinds of extensions, with stacking.
}

\subsection{Experiments on train+aug+val - test data}
\begin{table}[t]
\centering
 \begin{tabular}{l|r|r}
 \hline
 Method & IoU & IoU after \emph{Dense CRF}\\ \hline
Basenet & 72.72 & 73.78 \\ \hline
QO & 73.41 & 75.13  \\ \hline
QO$^s$ & 73.20 & 75.41  \\ \hline
QO$^{mres}$ & 73.86 & 75.46\\ \hline
  \end{tabular}
  \vspace{2mm}
\caption{Performance of our methods on the VOC PASCAL 2012 Image Segmentation Benchmark. Our baseline network (Basenet) is a variant of Deeplab-LargeFOV~\cite{deeplab1} network.
In this table, we demonstrate systematic improvements in performance upon the introduction of our Quadratic Optimization (QO), and multi-resolution (QO$^{mres}$) approaches. DenseCRF post-processing gives a consistent boost in performance.}
\label{table:results}
\end{table}
In this set of experiments, we train our methods on the \emph{train+aug+val} images, and evaluate them on the \emph{test} images.
The image resolutions and $\lambda$ values are the same as those in Sec.~\ref{subsec:prelim}. In these experiments, we also use the Dense CRF post processing as in \cite{deeplab1,deeplab2}.
Our results are tabulated in Tables \ref{table:results} and \ref{table:allresults}. 
We first compare our methods QO, QO$^s$ and QO$^{mres}$ with the basenet, where the relative improvements can be most clearly demonstrated.
Our multi-resolution network outperforms the basenet and other QO networks. We achieve a further boost in performance upon using the Dense CRF post processing strategy, consistently for all methods. 
We observe that our method yields an improvement that is entirely complementary to the improvement obtained by combining with Dense-CRF.

We also compare our results to previously published benchmarks in Table \ref{table:allresults}.
When benchmarking against directly comparable techniques, we observe that even though we do not use end-to-end training for the CRF module stacked on top of our QO network, our 
 method outperforms the previous state of the art CRF-RNN system of \cite{crfrnn} by a margin of $~0.8\%$. We anticipate further improvements by integrating end-to-end CRF training with our QO. 
In Table \ref{table:allresults}, we compare our methods to previously published, directly comparable methods,
namely those that use a variant of the VGG~\cite{vgg} network, are trained in an end-to-end fashion, 
and use structured prediction in a fully-convolutional framework.
\begin{table}[tbh]
\centering
 \begin{tabular}{l|r}
 \hline
 Method & mean IoU (\%) \\ \hline
Deeplab-Cross-Joint~\cite{deeplab2} & 73.9 \\ \hline
CRFRNN~\cite{crfrnn} & 74.7 \\ \hline
Basenet & 73.8 \\ \hline
QO & 75.1  \\ \hline
QO$^s$ & 75.4 \\ \hline
QO$^{mres}$ & 75.5 \\ \hline
  \end{tabular}
  \vspace{2mm}
  \caption{Comparison of our method with directly comparable previously published approaches on the VOC PASCAL 2012 image segmentation benchmark.}
  \label{table:allresults}
\end{table}

\subsection{Experiments with Deeplab-V2 Resnet-101}
\begin{figure*}[tbh]
\begin{center}
\includegraphics[width=0.23\linewidth]{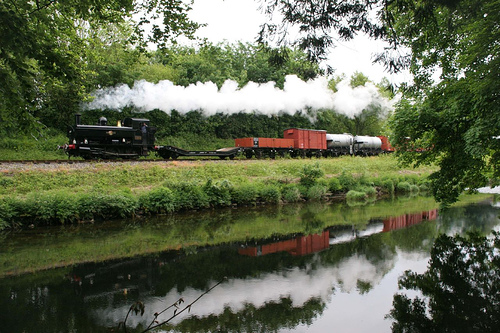}
\includegraphics[width=0.23\linewidth]{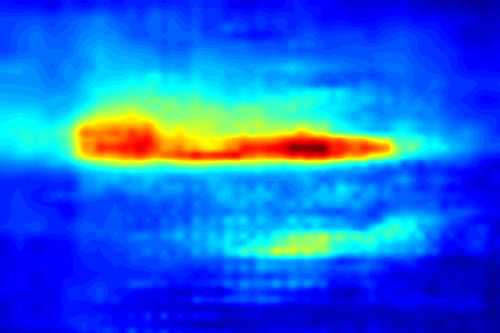}
\includegraphics[width=0.23\linewidth]{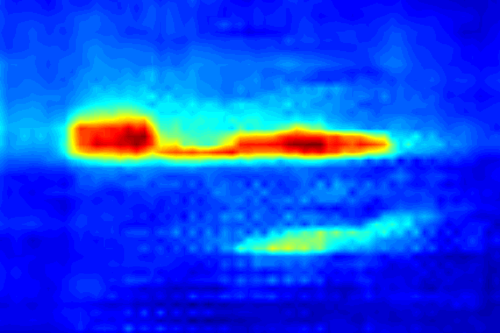}
\includegraphics[width=0.23\linewidth]{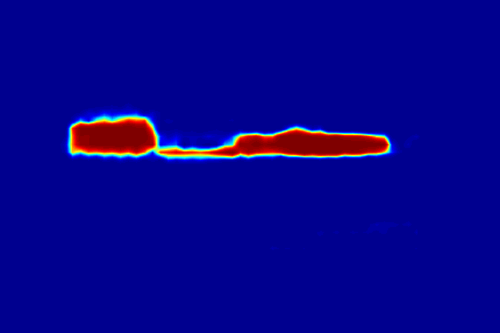}\\
\includegraphics[width=0.23\linewidth]{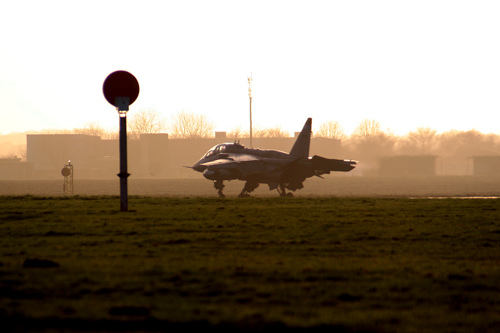}
\includegraphics[width=0.23\linewidth]{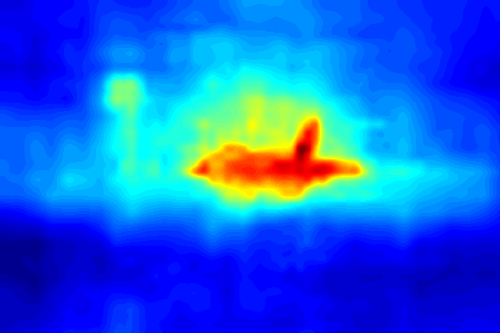}
\includegraphics[width=0.23\linewidth]{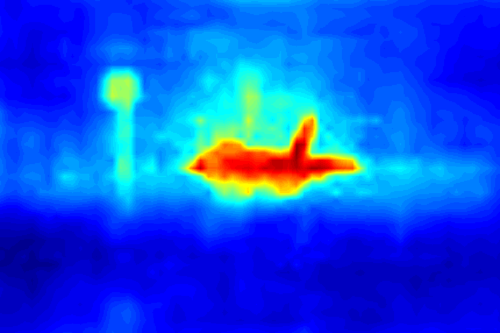}
\includegraphics[width=0.23\linewidth]{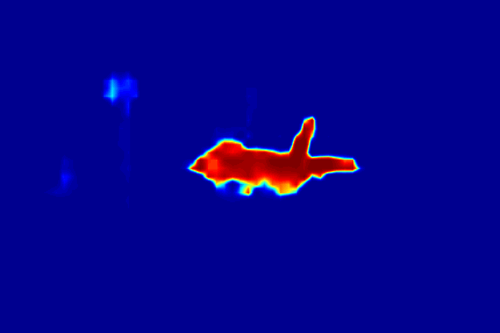}\\
\includegraphics[width=0.23\linewidth]{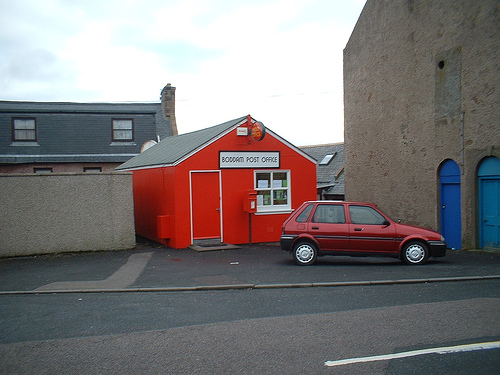}
\includegraphics[width=0.23\linewidth]{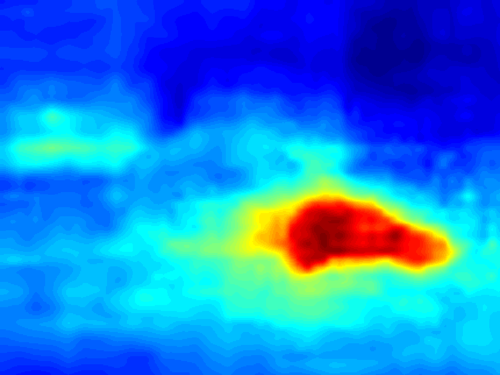}
\includegraphics[width=0.23\linewidth]{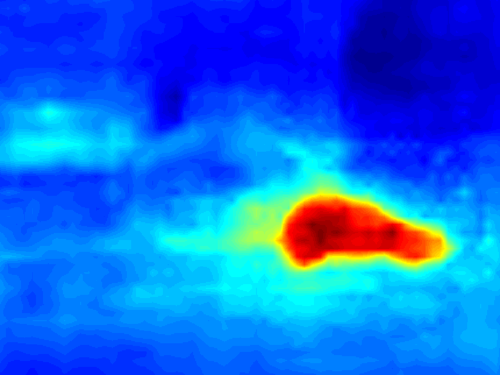}
\includegraphics[width=0.23\linewidth]{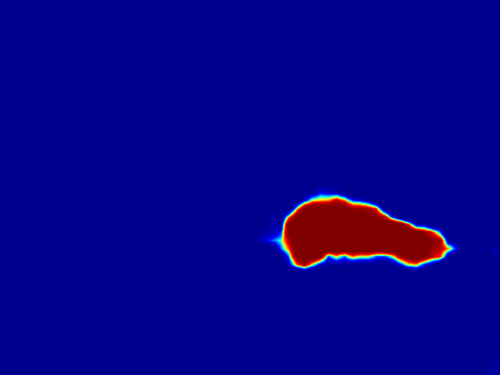}\\
\includegraphics[width=0.23\linewidth]{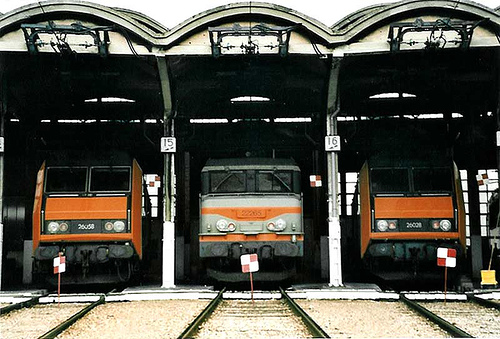}
\includegraphics[width=0.23\linewidth]{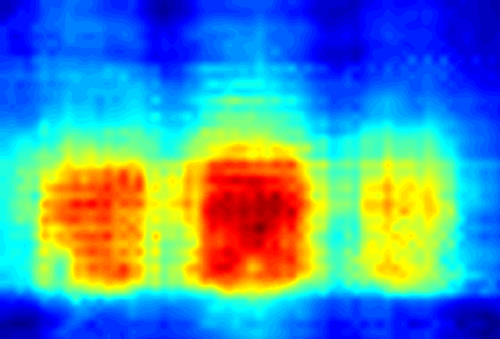}
\includegraphics[width=0.23\linewidth]{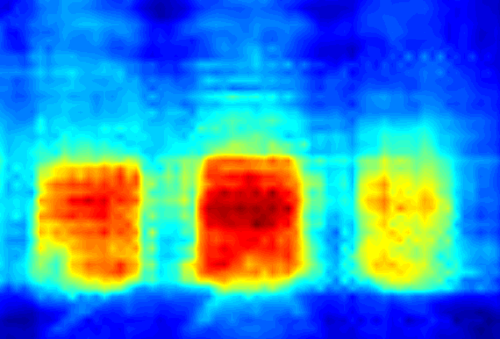}
\includegraphics[width=0.23\linewidth]{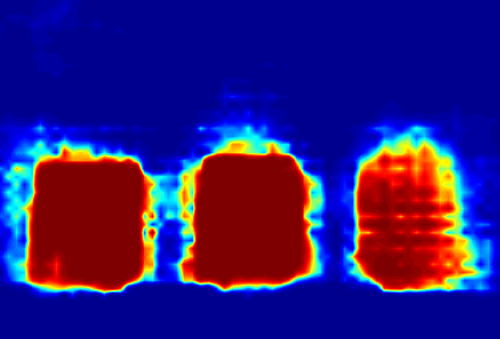}\\
\end{center}
\hspace{5mm} \textbf{(a) image} \hspace{15mm} \textbf{(b) unary} \hspace{15mm} \textbf{(c) output} \hspace{15mm} \textbf{(d) probability}\\
\caption{Qualitative results when our Potts type pairwise terms are used in combination with the deeplab-V2 Resnet-101 network. Column (a) shows the input image,
(b) shows the heatmap of the unary scores, (c) shows the heatmap of the scores after inference, and (d) shows the softmax probabilities. We notice that the object
boundaries are significantly finer after incorporating cues from the pairwise terms.}
\label{fig:visualresnet}
\end{figure*}

In this section we use our Potts-type model alongside the deeplab-v2~\cite{deeplabv2} Resnet-101 network.
This network is a $3$ branch multi-resolution version of the Resnet-101 network from \cite{resnet}. It processes
the input image at $3$ resolutions, with scaling factors of $0.5, 0.75$, and $1.0$, and then combines the network
responses at the different resolutions by upsampling the responses at the lower scales to the original scale, and
 taking an element-wise maximum of the responses corresponding to each pixel. We learn Potts type shared
 pairwise terms, and these pairwise terms are drawn from a parallel Resnet-101 network which has layers through
 \texttt{conv-1} to \texttt{res5c}, and processes the input image at the original scale. Table \ref{table:resnet} reports
 quantitative results on the PASCAL VOC 2012 test set. We show some qualitative results in Fig.~\ref{fig:visualresnet}.
It can be seen that our method refines the object boundaries, leading to a better segmentation performance.
 
 \begin{table}[tbh]
\centering
 \begin{tabular}{l|r}
 \hline
 Method & mean IoU (\%) \\ \hline
Deeplab-v2 + CRF ~\cite{deeplabv2} & 79.7 \\ \hline
QO$^s$ & 79.5 \\ \hline
QO$^s$ + CRF & 80.2 \\ \hline
\end{tabular}
  \vspace{2mm}
\caption{Performance of our Potts type pairwise terms on the VOC PASCAL 2012 test set with the deeplab-v2 Resnet-101 network.}
\label{table:resnet}
\end{table}

\section{Conclusions and Future Work}
In this work we propose a quadratic optimization method for deep networks which
can be used for predicting continuous vector-valued variables. The inference is efficient
and exact and can be solved in $0.02$ seconds on the GPU for each image in the general setting, and $0.003$ seconds for the Potts-type pairwise case using the conjugate gradient method.
We propose a deep-learning framework which learns features and model parameters simultaneously in an end-to-end FCN training algorithm.
Our implementation is fully GPU based, and implemented using the \emph{Caffe} library.
Our experimental results indicate that using pairwise terms boosts performance of the network
on the task of image segmentation, and our results are competitive with the state of the art methods
on the VOC 2012 benchmark, while being substantially simpler. While in this work we focused on simple $4-12$ connected neighbourhoods, we would like to experiment with fully connected graphical
models. Secondly, while we empirically verified that setting a constant $\lambda$ parameter brought about positive-definiteness, we are now exploring approaches to ensure this constraint in a general case. We intend to exploit our approach for solving other regression and classification tasks as in \cite{eigen2015predicting,ubernet}.
\\[1mm]

\noindent \textbf{Acknowledgements}
This work has been funded by the EU Projects MOBOT FP7-ICT-2011-600796 and I-SUPPORT 643666 \#2020.

\begin{figure*}
 \centering
\includegraphics[width=0.18\textwidth]{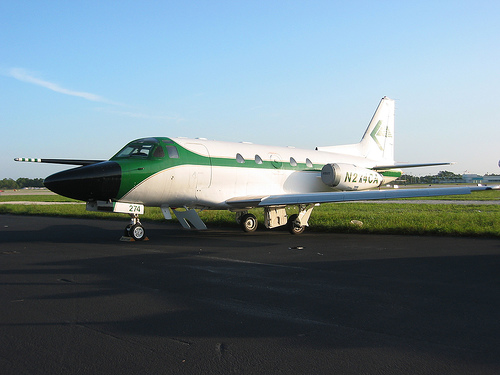}
\includegraphics[width=0.18\textwidth]{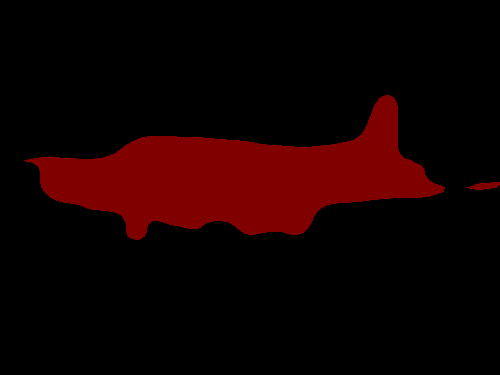}
\includegraphics[width=0.18\textwidth]{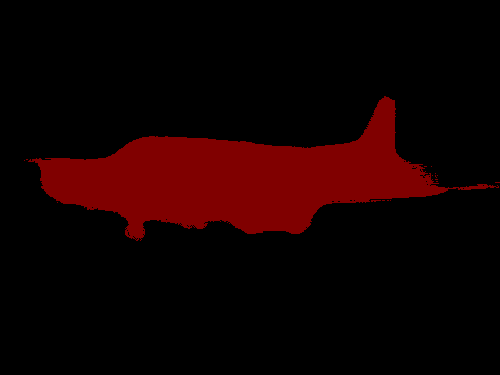}
\includegraphics[width=0.18\textwidth]{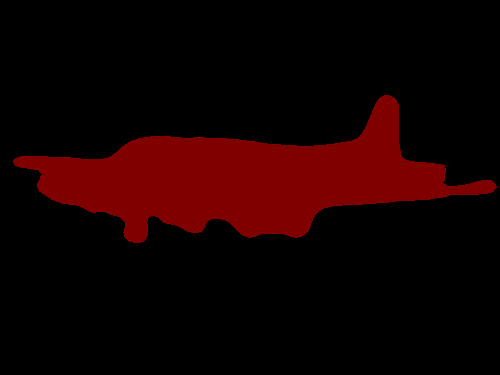}
\includegraphics[width=0.18\textwidth]{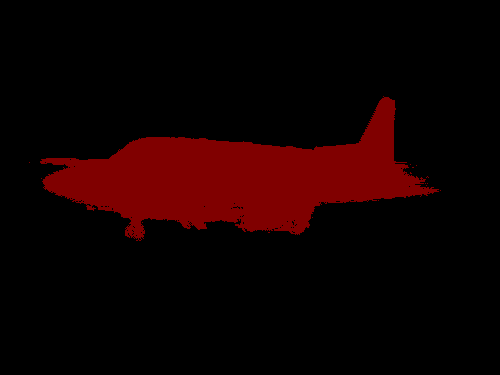}\\
\includegraphics[width=0.18\textwidth]{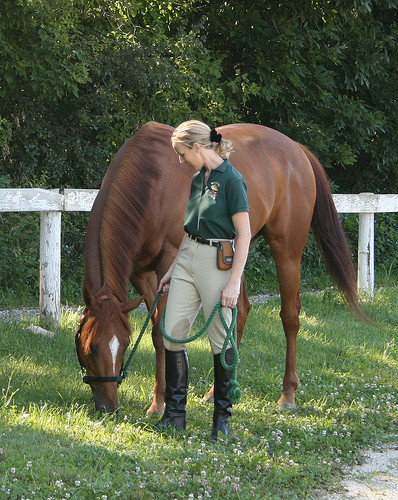}
\includegraphics[width=0.18\textwidth]{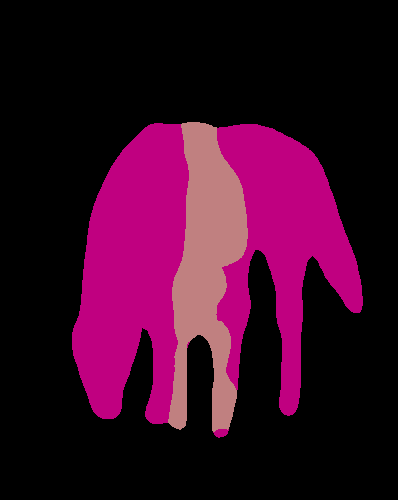}
\includegraphics[width=0.18\textwidth]{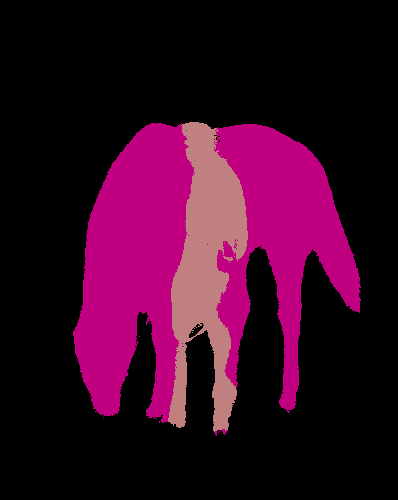}
\includegraphics[width=0.18\textwidth]{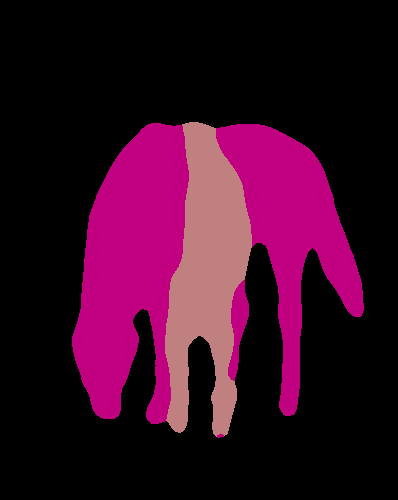}
\includegraphics[width=0.18\textwidth]{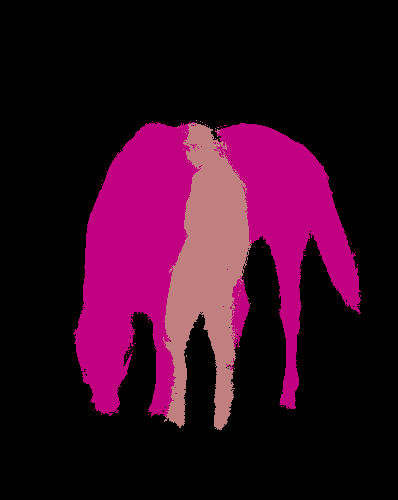}\\
\includegraphics[width=0.18\textwidth]{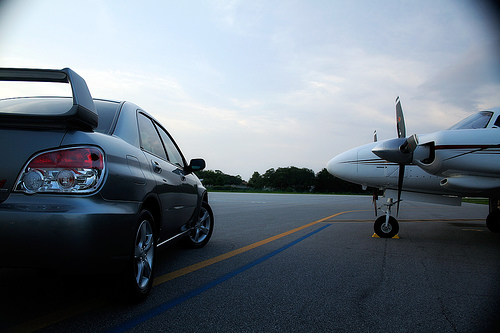}
\includegraphics[width=0.18\textwidth]{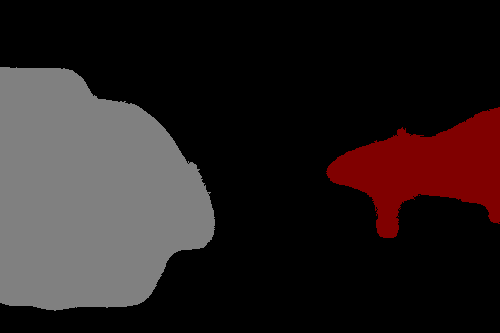}
\includegraphics[width=0.18\textwidth]{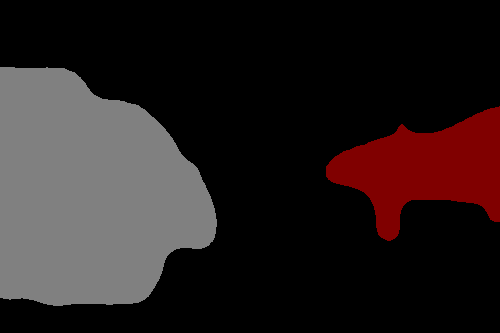}
\includegraphics[width=0.18\textwidth]{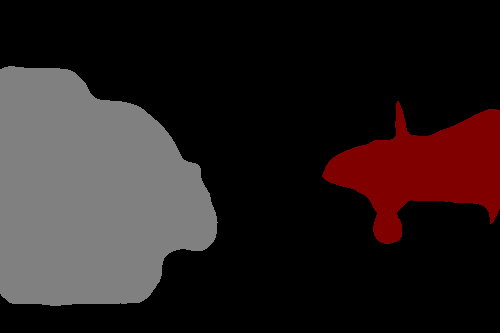}
\includegraphics[width=0.18\textwidth]{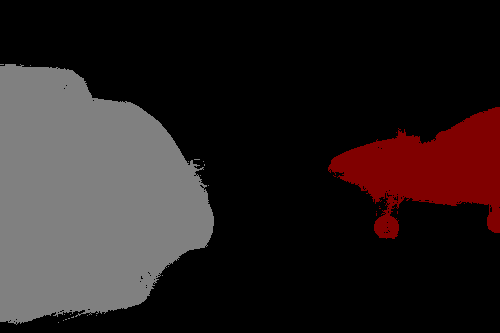}\\
\includegraphics[width=0.18\textwidth]{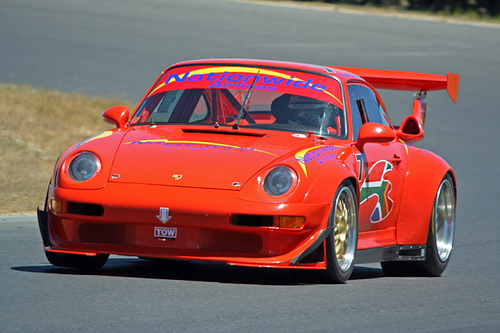}
\includegraphics[width=0.18\textwidth]{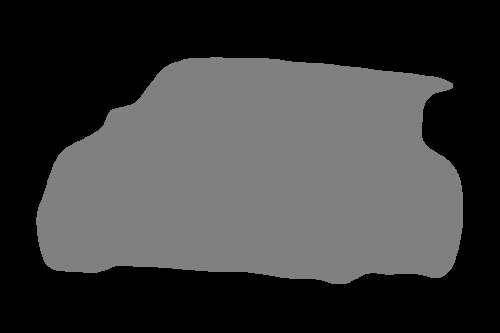}
\includegraphics[width=0.18\textwidth]{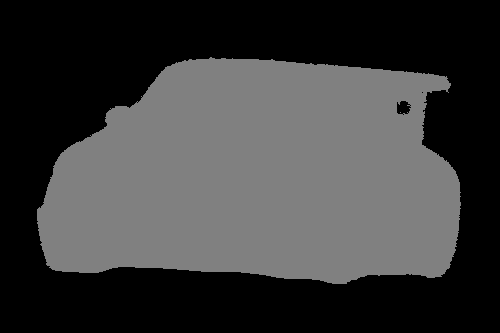}
\includegraphics[width=0.18\textwidth]{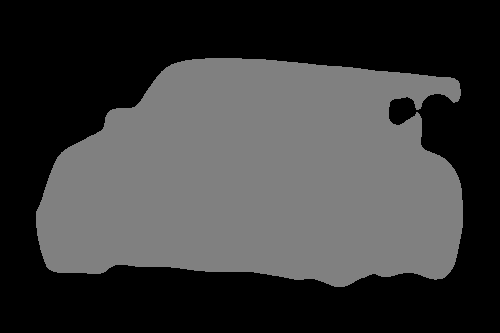}
\includegraphics[width=0.18\textwidth]{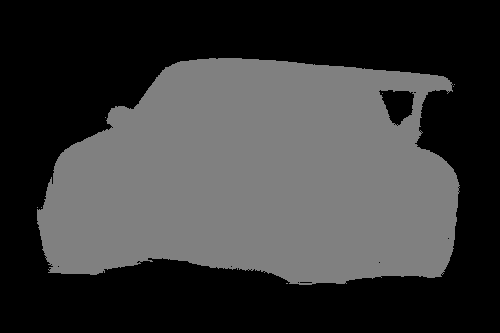}\\
\includegraphics[width=0.18\textwidth]{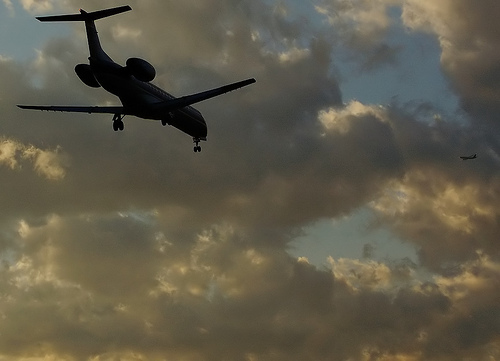}
\includegraphics[width=0.18\textwidth]{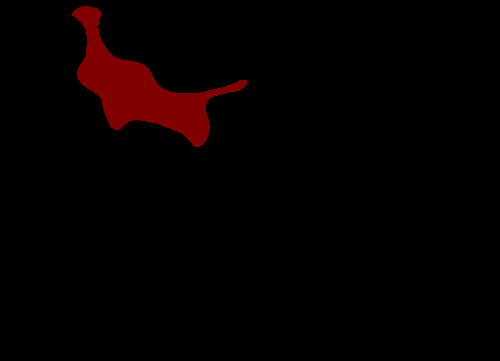}
\includegraphics[width=0.18\textwidth]{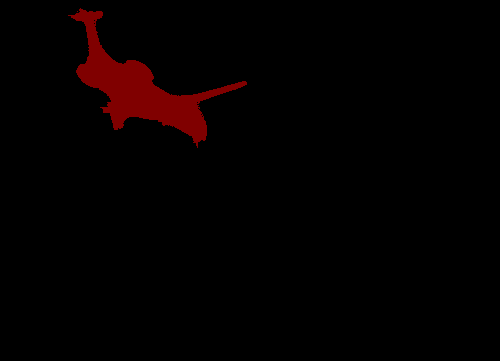}
\includegraphics[width=0.18\textwidth]{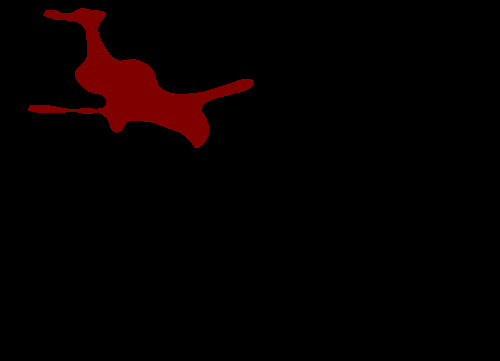}
\includegraphics[width=0.18\textwidth]{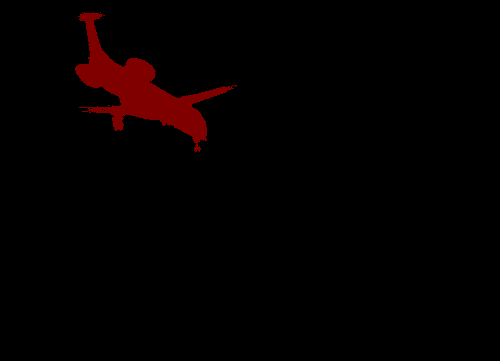}\\
\includegraphics[width=0.18\textwidth]{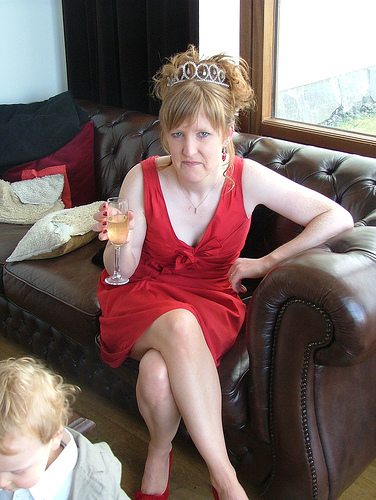}
\includegraphics[width=0.18\textwidth]{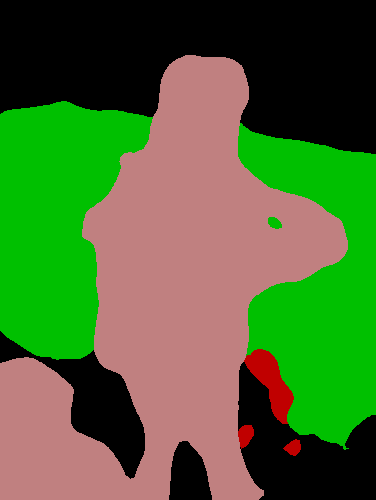}
\includegraphics[width=0.18\textwidth]{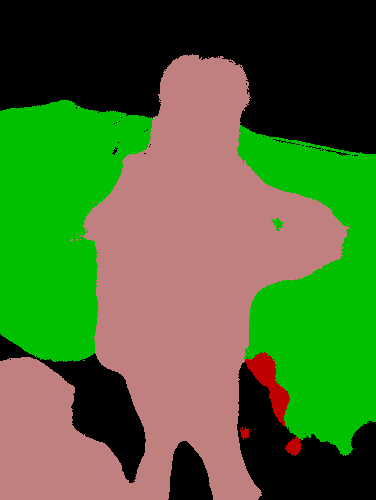}
\includegraphics[width=0.18\textwidth]{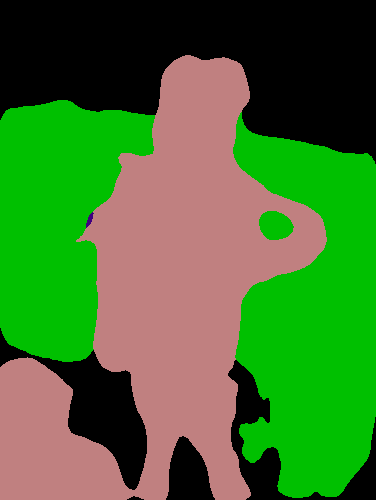}
\includegraphics[width=0.18\textwidth]{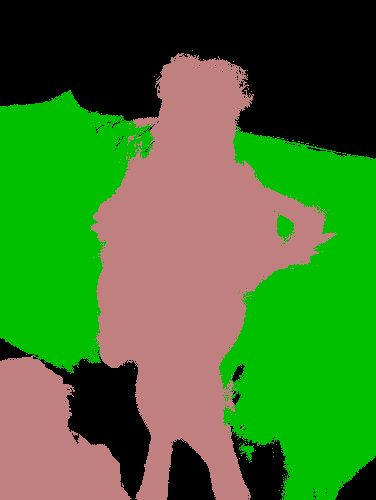}\\
\hspace{0mm} (a) Image \hspace{10mm} (b) Basenet \hspace{2mm} (c) Basenet + DCRF \hspace{3mm}  (d)  QO$^{mres}$ \hspace{0mm} (e) QO$^{mres}$ + DCRF \hspace{10mm}
\caption{Visual results on the VOC PASCAL 2012 test set. The first column shows the colour image, the second column shows the basenet predicted segmentation,
the third column shows the basenet output after Dense CRF post processing. The fourth column shows the
 QO$^{mres}$ predicted segmentation, and the final column shows the QO$^{mres}$ output after Dense CRF post processing.
It can be seen that our multi-resolution
network captures the finer details better than the basenet: the tail of the airplane in the first image, the person's body in the second image, the aircraft fan in the third image, the road between the
car's tail in the fourth image, and the wings of the aircraft in the final image, all indicate this. While Dense CRF post-processing quantitatively improves performance, it tends to miss very fine details.}
\label{fig:visual}
\end{figure*}

{\small
\bibliographystyle{splncs}
\bibliography{egbib}
}
\end{document}